\documentclass[sigconf]{acmart}

\AtBeginDocument{%
  \providecommand\BibTeX{{%
    \normalfont B\kern-0.5em{\scshape i\kern-0.25em b}\kern-0.8em\TeX}}}
\usepackage{times}
\usepackage{soul}
\usepackage{url}
\usepackage{graphicx}
\usepackage{amsthm}
\usepackage{booktabs}
\usepackage{algorithm}
\usepackage{algorithmic}
\usepackage[switch]{lineno}
\usepackage{xcolor}
\usepackage{longtable}
\usepackage{float}

\newcommand{\fyq}[1]{\textcolor{black}{#1}}
\newcommand{\zlh}[1]{\textcolor{black}{#1}}
\begin{document}

\title[Prompt as Free Lunch: Enhancing Diversity in CD-FSL through Text-Guided Prompt Tuning]{Prompt as Free Lunch: Enhancing Diversity in Source-Free Cross-domain Few-shot Learning through Semantic-Guided Prompting}
\author{Linhai Zhuo}
\email{linhaizhuo@fzu.edu.cn}
\affiliation{%
  \institution{College of Computer and Data Science, Fuzhou University}
  \city{Fuzhou}
  \country{China}}

\author{Zheng Wang}
\email{zhengwang@zjut.edu.cn}
\affiliation{%
  \institution{Zhejiang University of Technology, College of Computer Science}
  \city{Zhejiang}
  \country{China}}

\author{Tianwen Qian}
\affiliation{%
  \institution{Key Laboratory of Data Science and Intelligent Computing, Hangzhou International Innovation Institute, Beihang University}
  \city{Zhejiang}
  \country{China}}

\author{Yuqian Fu}
\affiliation{%
\institution{INSAIT, Sofia University}
\country{Bulgaria}}

\begin{abstract}
 The source-free cross-domain few-shot learning (CD-FSL) task aims to transfer pretrained models to target domains utilizing minimal samples, eliminating the need for source domain data. Addressing this issue requires models to have robust generalization abilities and strong feature representation, aligning with the characteristics of large-scale pretrained models. However, large-scale models tend to lose representational ability in cross-domain scenarios due to limited sample diversity. \zlh{Given the abundant diversity provided by semantic modality, this paper utilize prompt as "free lunch" to enhance the diversity and leverages textual modality to guide the prompt generating}.
 Specifically, we propose the SeGD-VPT framework, which is divided into two phases. The first step aims to increase feature diversity by adding diversity prompts to each support sample, thereby generating varying input and enhancing sample diversity. Furthermore, we use diversity descriptions of classes to guide semantically meaningful learning of diversity prompts, proposing random combinations and selections of texts to increase textual diversity. Additionally, deep prompt tuning is introduced to enhance the model's transfer capability. After training of the first step, support samples with different diversity prompts are input into the CLIP backbone to generate enhanced features. After generation, the second phase trains classifiers using the generated features. Extensive experimental results across several benchmarks verify our method is comparable to SOTA source-utilized models and attain the best performance under the source-free CD-FSL setting.
\end{abstract}




\maketitle
\section{Introduction}
Cross-domain few-shot learning (CD-FSL) marks a notable advance in machine learning, addressing the challenge of applying few-shot learning (FSL) principles across diverse data domains for a more realistic setting. It typically involves transferring models pretrained on source datasets like mini-ImageNet \cite{vinyals2016matching} to different target datasets, such as ChestX \cite{wang2017chestx}, with the main challenge being the substantial variation in data distributions, namely domain gap \cite{FWT}. Effective bridging this gap requires both robust generalization abilities and strong feature representation of the model \cite{wang2021,fu2022wave}. Given their proven strengths in feature generalization and representation   \cite{wang2023large}, the integration of large-scale pretrained models significantly enhances performance in CD-FSL tasks, as indicated in \cite{hu2022pushing,fu2023styleadv}.\par

However, with the introduction of such well-pretrained models, a question arises: is the source dataset still necessary? Intuitively, adopting a source-free CD-FSL setting has two significant advantages: 1) it resonates with the original purpose of few-shot learning by eliminating the need for collecting source domain data; 2) directly fine-tuning on the target domain avoids the influence of an external domain (like mini-ImageNet) on the large-scale pretrained models, thus enabling a more focused exploration of these models' cross-domain capabilities. 
\fyq{Consequently, in this paper, we opt for the source-free CD-FSL.}
\par
\begin{figure}[t]
  \centering
    \vspace{-0.1in}
  \includegraphics[width=\linewidth]{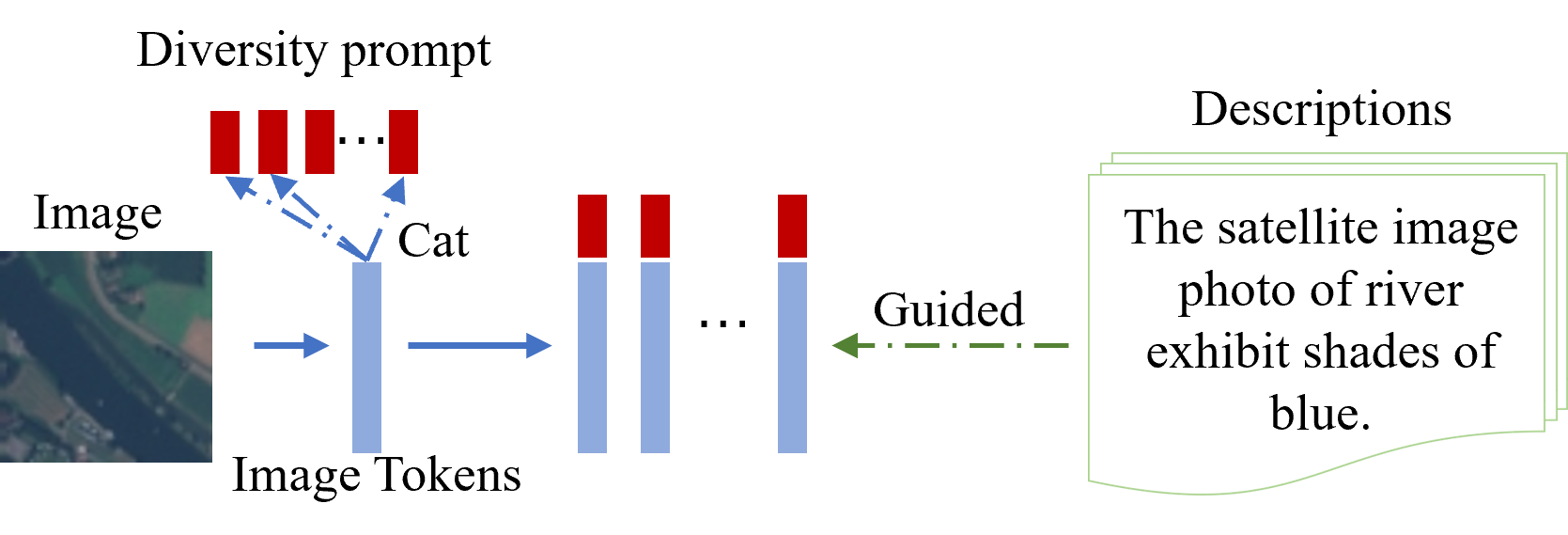}
  \caption{
  Illustration of SeGD-VPT: 1) diversity prompts are integrated into images to boost visual diversity; 2) class descriptions are employed to guide the learning of diversity prompts. }
  \label{fig:tease}
  \vspace{-0.25in}
\end{figure}

In exploring the application of the large scale model in source-free CD-FSL scenarios, two key challenges emerge: 
1) A notable mismatch exists between the data distributions of target domains and pretrained models. For example, the ChestX dataset, composed entirely of X-ray images, significantly impacts the visual encoder. This leads to challenges in transferring the encoder to recognize the unique features of target domain images.
2) Furthermore, the inherent scarcity of diversity in CD-FSL tasks becomes even more challenging for large-scale pretrained models with extensive parameters. The limited data, coupled with the models' complexity, can exacerbate the difficulty in accurately mirroring the target domain's data distribution. This not only compromises the representational capacity but also significantly impairs the models' ability to generalize effectively to the target domain, highlighting a critical hurdle in achieving high performance across diverse domains. As far as we know, there are only three works proposed in CD-FSL task to solve the above problems \fyq{based on the large-scale pretrained models.} For the first challenge, \citet{hu2022pushing} use meta-learning and \citet{ma2023prod} introduce domain-specific prompts with multi-domain data to boost generalization. For the second, \citet{fu2023styleadv} employ adversarial training to create hard samples. However, these approaches are sensitive to learning rate adjustments and require source data.
\par

In this paper, we argue that prompts can serve as a free lunch to enhance sample diversity, guided by the textual modality. As shown in Figure \ref{fig:tease}, by cascading different prompts, diverse inputs can be generated. Additionally, to ensure the added prompts are meaningful, we refer to the textual modality to guide the learning of the prompts. Given that textual modality inherently offers richer diversity due to the varied perspectives present in textual descriptions, this diversity can make the added prompts semantically richer. Furthermore, textual descriptions exhibit better consistency across different domains, maintaining uniformity and showing less susceptibility to changes in context or appearance. Consequently, this makes the textual modality highly suitable for guiding prompt learning in cross-domain scenarios.
\fyq{Therefore, in this paper, we mainly focus on using prompts to expand the data distribution and leverage the textual modality as guidance to generate prompts for learning. With the additional textual guided prompted features, we are further able to facilitate the transferability of large-scale pretrained models.}

Specifically, this paper introduces \textbf{Semantic Guided Diversity Visual Prompt Tuning (SeGD-VPT)} framework to tackle the two challenges in source-free CD-FSL tasks, by incorporating cross-modality large scale pretrained model CLIP and visual prompt tuning. As illustrated in Figure \ref{fig:tease}, this method aims to improve the diversity of the input by adding different diversity prompts to the support sample, guiding prompt generation through semantic descriptions, and concurrently conducting model transfer through such cross-domain generation tasks. 

Concretely, this paper tackles the issue of limited sample diversity and capitalizes on the advantages of prompting and textual modality in CD-FSL from three perspectives: \textbf{visual},\textbf{ textual}, and \textbf{cross-modality}: 1) From a \textbf{visual} standpoint, this paper proposes to add learnable diversity prompt tokens at the image input layer to increase the diversity of the CLIP visual encoder's input; 2) For the \textbf{textual} modality, various descriptions of categories are collected from the web as describe prompts and fed into CLIP's text encoder to extract features. By randomly combining these features, an abundance of diversity semantic feature are generated. To ensure these features' consistency with the classification task, contrastive learning is used to align the diversity semantic features with class prompts (a [Domain] photo of [Class]); 
3) From a \textbf{cross-modality} perspective, diversity semantic features are used to guide the training of diversity prompts, with Target Supervised Contrastive Learning \cite{li2022targeted}, making it more contextually rich and relevant to the class. Additionally, randomly 
\fyq{selection} is applied to further enhance diversity. Moreover, deep prompts are employed to minimize learnable parameters, thereby mitigating the risk of overfitting. Finally, once the diversity prompts are well-trained, they are fed to the visual encoder to produce prompt visual features, which are then utilized to train the classifier. Our experiments on the BSCD benchmarks dataset \cite{guo2020broader} demonstrate that the proposed SeGD-VPT framework achieves accuracy comparable to SOTA models trained on the source dataset and attains the best performance under the source-free CD-FSL setting.\par
Overall, our contributions can be summarized as follows: 
\begin{itemize}
    \item[1)] Semantic Guided Diversity Visual Prompt Tuning (SeGD-VPT) framework is introduced for source-free CD-FSL, enhancing data diversity and the CLIP model's transferability through diversity semantic feature guided diversity prompts. 
\item [2)] We show that the semantic features, specifically diverse features containing varied descriptions, help facilitate domain transfer \fyq{effectively.} 
    \item[3)] The effectiveness of our method is proven by experiments on four datasets: ChestX, ISIC, EuroSAT, and CropDisease.
\end{itemize}

\section{Related Works}

\subsection{Cross-Domain Few-Shot Learning}
Cross-Domain Few-Shot Learning (CD-FSL) focuses on recognizing images from different distributions compared to the training set. Traditional methods in CD-FSL have aimed to enhance the generalization capabilities of models. 
\fyq{For example, through Gaussian noise in FWT~\cite{FWT}, explanation guidance in LRP~\cite{sun2021explanation}, adversarial training in ATA~\cite{wang2021} and AFA~\cite{hu2022adversarial}, style augmentation in wave-SAN\cite{fu2022wave}  and StyleAdv~\cite{fu2023styleadv}.}
\fyq{Some research has focused on adapting deep learning models to the target domain through finetuning support data from target domains. Typical methods include Fine-tune~\cite{guo2020broader}, NSAE~\cite{liang2021boosting}. BSR~\cite{liu2020feature}, DARA~\cite{zhao2023dual}. ATA and StyleAdv also apply finetuning on the meta-trained models to obtain better results. }
\fyq{Methods e.g., STARTUP~\cite{startup}, Meta-FDMixup~\cite{metafu}, DDN~\cite{ddn}, CLD~\cite{zheng2023cross}, TGDM~\cite{zhuo2022tgdm}, ME-D2N~\cite{fu2022me} explore CD-FSL by integrating additional target domain data into the training stage.}
However, the introduction of large pretrained models in CD-FSL is relatively less explored, with notable exceptions like PMF~\cite{hu2022pushing}, StyleAdv~\cite{fu2023styleadv}, and ProD~\cite{ma2023prod}, which examine the cross-domain capabilities of large-scale pretrained models in conjunction with few-shot datasets. 
\fyq{Among them, both PMF and StyleAdv require training on the single source, while ProD employs visual prompt tuning with a "leave one out" approach, requiring data from both the source and multiple other domains. By contrast, this paper investigates the source-free setting which is much more challenging.}
\fyq{VDB~\cite{yazdanpanah2022visual} and IM-DCL~\cite{xu2024enhancing} which both study the source-free CD-FSL may two most related works to us. However, technically, they tackle from the perspective of batch normalization and  Information Maximization respectively. While} this paper focuses on creating more diversity with limited target domain data to avoid collapse while utilizing VPT to transfer large-scale models to the target domain.

\subsection{Prompt Tuning}
Prompting \cite{liu2023pre} initially means appending manually chosen language instructions to input text, allowing pre-trained language models to 'understand' downstream tasks. Recent approaches treat prompts as task-specific continuous tokens/vectors, optimizing them via gradients during fine-tuning, known as Prompt Tuning \cite{li2021prefix,lester2021power,liu2021p}. Recently, Prompt Tuning has also been employed in vision and multimodal tasks \cite{qian2023decouple,liu2023generating,liu2022prompt,auty2023learning,yan2023prompt}. Its ability to effectively reduce learnable parameters makes it popular in few-shot learning, i.e. RPO \cite{lee2023read}, VPPT \cite{song2023vppt}, RePrompt \cite{rong2023retrieval} and LoCoOP \cite{miyai2023locoop}. Besides, prompt tuning is also utilized to domain generation task to transfer model to different domain, i.e. Stylip \cite{bose2024stylip}, DPL \cite{zhang2023domain}, CSVPT \cite{li2022learning} and Promptstyler \cite{cho2023promptstyler}. Notably, Promptstyler proposes to generate diversity style in language space. In this paper, we create diversity in visual-language space through diversity prompt while transfer large model to target domain in source-free CD-FSL task. 

\subsection{Few-Shot Learning with Large Scale Pretrained Model}
Integrating large-scale pretrained models has significantly advanced FSL task, bolstering adaptability with limited data. Notably, CLIP-Adapter \cite{gao2021clip} incorporates FC layer with residual connections to CLIP, while Tip-Adapter \cite{zhang2021tip} adopts a parameter-free strategy using a cache model. Efforts to integrate diverse models like GPT-3, DALL-E, and CLIP \cite{zhang2023prompt}, along with meta distillation \cite{wu2023few} and finetuning schedulers \cite{goyal2023finetune,chen2023knowledge}, exemplify this progression. This paper focus the domain
transfer ability of large-scale model through visual prompt tuning.


\section{Method}
\subsection{Preliminaries}
Assuming a target domain dataset $D_T$, an episode $E = \left\{ \left( S, Q \right), Y \right\}$ is randomly sampled for meta-testing. This meta-testing is formulated as an N-way K-shot problem. Specifically, for each episode $E$ from $D_T$, $N$ classes with $K$ labeled images are sampled to form the support set $S$, and the same $N$ classes with $M$ different images comprise the query set $Q$. The set of labels for these $N$ classes is denoted as $Y = \left\{{c_i}\right\}_{i=1}^{N}$. The support set $S$ is utilized for training the model, while the query set $Q$ is used to evaluate accuracy.
\begin{figure*}[ht]
  \centering
  \includegraphics[width=0.92\linewidth]{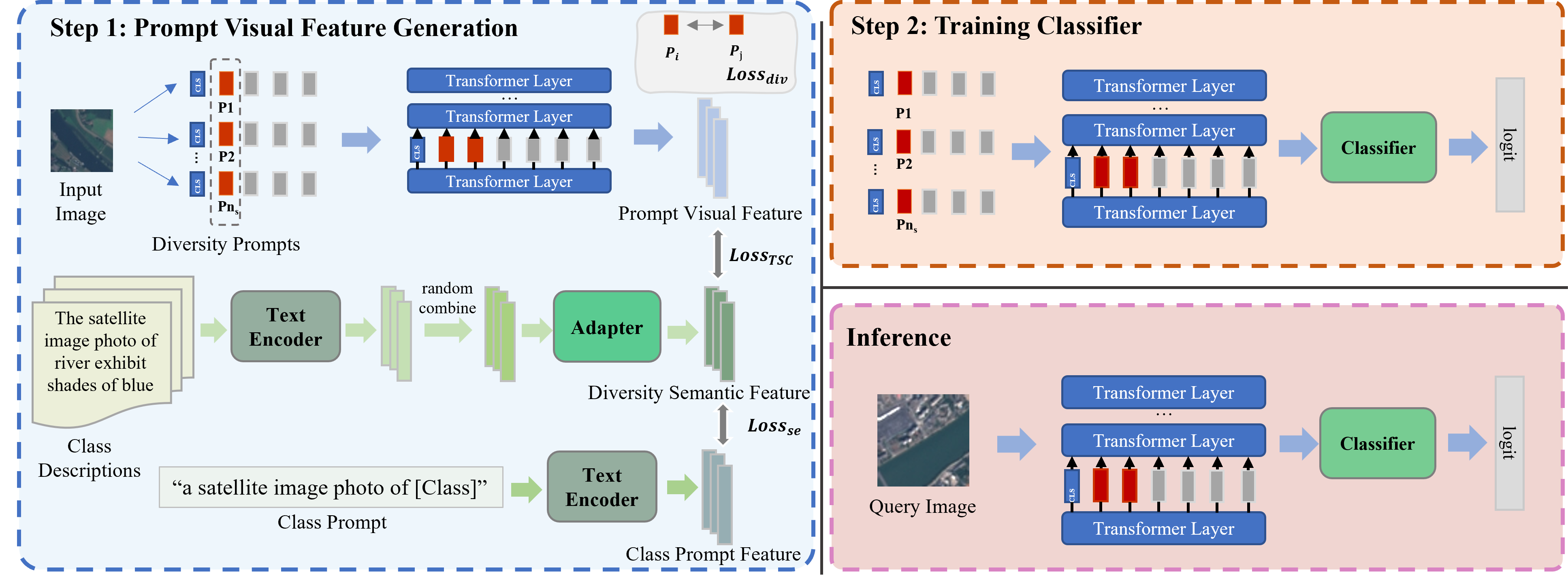}
  \caption{Pipeline of the proposed SeGD-VPT framework. The SeGD-VPT employs a two-step training process. Step 1 aims to generate prompt visual features guided by semantics to enhance data diversity while transferring the CLIP model. In Step 2, the classifier is trained using the prompt visual features generated in Step 1. }
  \label{fig:framework}
\end{figure*}
\subsection{Overview}
The overall framework of the proposed method is shown in Figure \ref{fig:framework}, and the pseudo-code of SeGD-VPT is described in Algorithm \ref{algorithm1}. Our method guides the learning of diversity prompt and deep prompt tokens within the visual-language space (e.g., CLIP latent space) using diversity semantic features. After training, images combined with diversity prompt are fed into transformer layers containing deep prompt tokens to extract features with more diversity. These features are then used to train a classifier. During inference, images are processed through the transformer with the deep prompts and subsequently classified by the classifier. Notably, CLIP serves as our large-scale cross-modality model to map both visual and textual input into the same hyperspace. And its visual and text encoders remain frozen throughout the training and inference processes.

\subsection{SeGD-VPT Model}
SeGD-VPT primarily aims to enhance the sample diversity in CD-FSL, thus averting the collapse in prompt tuning and enabling a more effective model.
This is achieved by strategically increasing sample diversity from three distinct aspects: visual, textual, and cross-modal. Each of the three modalities will be introduced separately, followed by a description of the overall training process.\par

\textbf{For the visual modality}, random augmentations (e.g., flipping, translation) are applied to each support images $n_v$ times, creating $N \times K \times n_v$ variants. These augmented images are then divided into $c$ segments and transformed into vision tokens through CLIP's token encoder. We concatenate diversity prompt tokens of length $l$ at the front of the image tokens for each augmented image, with each diversity prompt being independent. Denoting this set of diversity prompts as $P_{div}$, the input to the vision transformer can be structured as $(S,P_{div})$ accordingly. Deep prompt tokens $P_{deep}$ are integrated into CLIP's visual encoder, for two main purposes: firstly, fine-tuning deep prompts while keeping other parameters fixed reduces the number of learnable parameters, helping to prevent overfitting in scenarios with limited samples. Secondly, since the diversity prompts are independent, this paper aims for the deep prompt tokens to learn shared knowledge like visual-textual mapping and classification capabilities, thus transferring the model's representational and classification abilities to the target domain. The visiual encoder with deep prompt tokens is denoted as $E_v(x;P_{deep})$, where $x$ is the input. $(S,P_{div})$ are fed into $E_v(x;P_{deep})$ to produce the visual modality's output features $F_{v}$, namely prompt visual feature. To ensure the diversity of inputs, prompt diversity loss $L_{div}$ is designed to prevent the convergence and collapse of the diversity prompts during learning. This loss function is to minimize the cosine similarity between different diversity prompts and is formulated as:
\begin{equation}
L_{div} = \frac{1}{N_v} \sum_{i=1}^{N_v} \frac{1}{N_v-1} \sum_{j=1, j \neq i}^{N_v} \left| \frac{P_{div}^i}{\big\| P_{div}^i \big\|_2} \cdot \frac{P_{div}^j}{\big\| P_{div}^j \big\|_2} \right|,
\label{eq:Ldiv}
\end{equation}

where $N_v$ is the total number of diversity prompt $P_{div}$, which is equal to $N\times K\times n_v$ .\par

\textbf{For textual modality}, class descriptions are collected from the target domain (e.g., via ChatGPT) and convert these into single-feature named describe prompts, with $n_s$ prompts per class. For example, in the EuroSAT dataset's 'river' category, descriptions include seasonal color changes (e.g. green in summer, yellow in autumn) and other broader characteristics. More details are in the supplementary material. These prompts are processed through CLIP's text encoder to extract $n_s$ describe text prompt features. To enhance the diversity of text features, text prompt features are augmented by $t_s$ times via random combinations, forming a set denoted by $F_{dp}$. This set $F_{dp}$ is subsequently processed through a learnable Adapter \cite{gao2021clip}, represented as $A_s(x)$. Eventually, $N_s = N \times t_s \times n_s$ output features $F_s$ are obtained, named diversity semantic features. Simultaneously, class prompts (e.g. a [Domain] photo of [Class]) are fed to text encoder without Adapter and obtain $N$ class prompt features $F_{cls}$. Semantic contrastive loss $L_{se}$ is designed to align diversity semantic features with corresponding class prompt features, ensuring classification consistency. The loss formula is detailed as follows:
\begin{align}
    &Sim_{i,j} = \left|\frac{A_s(F_{s}^i)}{\left\| A_s(\begin{aligned}F\end{aligned}_{s}^i) \right\|_2} \cdot \frac{F_{cls}^j}{\left\| \begin{aligned}F\end{aligned}_{cls}^j \right\|_2 }\right| \\
    &L_{se} = -\frac{1}{N_s} \sum_{i=1}^{N_s} log(\frac{exp(Sim_{i,y})}{ {\textstyle \sum_{z=1}^{N}}exp(Sim_{i,z}) } ),
    \label{eqL:Lsc}
\end{align}
where $y$ is the label of $F_s^i$, $Sim_{i,j}$ is an intermediate variable that calculates the cosine similarity between \( i^{th} \) diversity semantic feature and \( j^{th} \) class prompt feature.\par

\textbf{From a cross-modality standpoint}, we align prompt visual features $F_v$ with nearest diversity semantic features $F_s$ in hyperspace while preserving diversity. This ensures the learning of diversity prompt more semantic meaningful. Additionally, since the task is trained on data from target domain, it facilitates the transfer of the CLIP model to target domain as well. The process is as follows: first, for each visual feature $v_i$ from $F_v$, top $c$ features  are selected from the same category in $F_s$ based on cosine similarity, filtering out unreasonable text feature combinations. To maintain diversity, $m$ text features $\tilde{F_s^i}$ from these $c$ ones are selected randomly according to gamma distribution. Target Supervised Contrastive Loss (TSC loss \cite{li2022targeted}) is adopted to align the visual feature with these $m$ selected text features. Specifically, the TSC loss $L_{TSC}$ is formulated as: 
\begin{equation}
  L_{TSC} = -\frac{1}{N_v}\sum_{i=1}^{N_v}\frac{1}{m} \sum_{j=1}^{m} log\frac{e^{v_i\cdot {^{}}  v_j^t /\tau } {}}{\sum_{v_s\in F_s}^{}e^{v_i\cdot v_s/\tau} },   
  \label{equ:TSC}
\end{equation}
where $v_s$ is the semantic feature belong to $F_s$ and $\tau$ is temperature parameter. Notably, as the limited visual features may not fully represent the feature distribution, we aim to prevent their influence on text features. Hence, $L_{TSC}$ is applied only to update learnable prompt tokens, i.e. $P_{div}$ and $P_{deep}$, not the text Adapter $A_s(x)$.\par

Overall, in this phase, the focus is on training $P_{div}$, $P_{deep}$ and $A_s(x)$ to than generate prompt visual features. The procedure unfolds as follows: initially, images are concatenated with diversity prompts and fed into  $E_v(x;P_{deep})$ to produce diversity prompt features. Next, $F_{dp}$ are fed to $A_s$ to obtain diversity semantic features. The diversity prompt features are then paired with corresponding diversity semantic features selected based on similarity and randomness. Then, $L_{div}$, $L_{se}$, and $L_{TSC}$ are computed as Equation \ref{equ:TSC}. The visual loss $L_v = L_{div} + L_{TSC}$ is employed to train $P_{div}$ and $P_{deep}$. Meanwhile, only $L_{se}$ is applied to train $A_s(x)$.
The training process is repeated for $T$ iterations. Post-training, the prompt visual feature $F_v$ are regenerated. These features display enhanced diversity compared to the originals, driven by textual features that concentrate around class centers, ensuring the requisite consistency for classification. 

\subsection{Classifier Training}
After acquiring $N_s$ prompt visual features from SeGD-VPT, these are input into a classifier $G_{cls}(x)$ composed of an Adapter and a fully connected layer. The classifier is trained using Arcface loss \cite{deng2019arcface} as classification loss $L_{cls}$, which is cosine similarity-based. This loss aims to narrow the distance between classifier weights and same-category features while widening the gap for features from different categories with an angular margin. Being similarity-based, Arcface loss is apt for visual-language hyperspace features.
\subsection{Inference}
During inference, both the trained classifier and the deep prompt-enhanced visual encoder are employed. For any given input image $x_i$, the visual encoder initially extracts its features. These features are then fed into the classifier, which calculates the probability for each class. The class with the highest probability is selected as the predicted category $C_{pred}$. Notably, the text modality including text encoder and text adapter are not utilized in this inference process. The complete formula for inference is outlined below:
\begin{equation}
    C_{pred} = argmax \ \ G_{cls}(E_v(x_i;P_{deep})).
\end{equation}

\section{Experiment}
\noindent\textbf{Datasets:} In this study, we do not utilize source domain dataset and finetune our model directly on target domain. Specifically, for the target domain, we utilize the BSCD dataset \cite{guo2020broader}, which amalgamates four distinct datasets: ChestX \cite{wang2017chestx}, ISIC \cite{tschandl2018ham10000,codella2019skin}, EuroSAT \cite{helber2019eurosat}, and CropDisease \cite{mohanty2016using}.\par
\noindent\textbf{Network Modules:} In our model, we adopt the Vit-base/16 network architecture as the principal feature extraction network, with parameters pre-trained from the DFN2B dataset \cite{ilharco_gabriel_2021_5143773}. The text adapter and the adapter in classifier are with the same parameters and architecture, comprising three FC layers with dimensions [512,128,512]. For deep prompts, the number of tokens per layer are set at 5. And for the diversity prompt, we standardize the total number of samples used across all tasks for fine-tuning (including both original support features and generated features) at 25. This means, for a 1-shot task, each image is paired with 24 diversity prompts; similarly, for 5-shot task, each image is paired with 4 diversity prompts. The SeGD-VPT in this experiment includes the classifier from Section 3.4.\par
\noindent\textbf{Implemental Details:} In our experimentation, we adopt both the 5-way 1-shot and 5-way 5-shot scenarios. We evaluate our network by using 15 query samples per class, randomly selecting 1000 tasks, and reporting the average results(\%). The length of diversity prompt $l$ is set to 1. For the fine-tuning phase, we conduct training across epochs $T$ from 40 or 55 or 60, utilizing the Adam optimizer with learning rates $lr$ of 0.001 and 0.0001. More training details are presented in supplementary materials. In the process of randomly selecting diversity semantic features, we employ a top-c strategy with $c$ and $m$ equals 300 and 100, respectively. And the random strategy follows Gamma (2.0, 75). The temperature parameter $\tau$ in Equation \ref{equ:TSC} is fixed at 0.07. All training and testing procedures are executed on NVIDIA 4090 or A100 graphics card.
\subsection{Comparison with the SOTAs}
We compare our SeGD-VPT framework against several most representative and competitive CD-FSL methods. totally 14 methods are used as competitor with different setting including different backbone, whether using source dataset and whether finetuning on target domain. 
\fyq{Concretly, the 14 methods include GNN \cite{gnn}, FWT \cite{FWT}, LRP \cite{sun2021explanation}, ATA~\cite{wang2021cross} , ATA-FT (formed by finetuning ATA), AFA~\cite{hu2022adversarial}, wave-SAN~\cite{fu2022wave}, StyleAdv~\cite{fu2023styleadv}, StyleAdv-FT (finetuned StyleAdv), DARA~\cite{zhao2023dual}, Fine-tune~\cite{guo2020broader}, NSAE~\cite{liang2021boosting}, BSR~\cite{liu2020feature}, PMF \cite{hu2022pushing}, VDB~\cite{yazdanpanah2022visual}, IM-DCL~\cite{xu2024enhancing} are introduced as our competitors. }
\fyq{Among them, GNN, FWT, LRP, ATA, AFA, waev-SAN, StyleAdv (RN10) all use the ResNet10 as backbone and perform the direct inference; ATA-FT, DARA, StyleAdv-FT (RN10) further include finetuning on target support images; PMF and StyleAdv-FT (ViT) build on DINO pretrained ViT and require finetuning; FN+VDB and IM-DCL are two source-free CD-FSL methods with different backbones. }
Those methods that use extra target training datasets, e.g., STARTUP~\cite{startup} and meta-FDMixup~\cite{metafu} are not considered. The comparison results are given in Table \ref{tab:main}. \par
Across the entirety of our results, our method significantly surpasses the established benchmarks in Cross-Domain Few-Shot Learning (CD-FSL), thereby establishing a new state-of-the-art. Notably, our SeGD-VPT model achieves average accuracy of 58.31\% and 66.76\% in the 5-way 1-shot and 5-way 5-shot settings, respectively. 

Beyond setting new accuracy records, our study uncovers several important findings:
1) Despite forgoing pre-training on the source domain, our approach not only surpasses methodologies that leverage such pre-training in terms of average performance and across the majority of datasets but also achieves state-of-the-art results. 
\fyq{We could observe this from the EuroSAT and CropDisease datasets. }
For instance, our method's performance exceeds the second-best by considerable margins in both the 1-shot and 5-shot tasks. This demonstrates the effectiveness of our strategy in directly adapting large-scale pre-trained models to various target domains with minimal samples, thereby eliminating the need for additional pre-training on the source domain. It also suggests the enhancement potential of leveraging large-scale pre-trained models.
\fyq{2) Compared to similar approaches that do not utilize source domain data, our model also demonstrates significant advantanges.  Concretly, our SeGD-VPT improves the VDB by 7.07\%, 2.02\% on 1-shot and 5-shot respectively. For the more competitive IM-DCL we also outperform by 2.40\% on 1-shot.}
These enhancements underscore the effectiveness of our proposed methodology and the utility of large-scale pre-trained models.
\fyq{We also note that the IM-DCL performs better than us on ChestX and ISIC datasets, however, we highlight that IM-DCL adopts a transductive setting which uses all the query images during the inference stage while we don't rely on that. }
3) Accuracy rates for methods based on fine-tuning, including our proposed SeGD-VTP, are generally higher than those that do not employ fine-tuning. This indicates that fine-tuning is an effective strategy for enhancing accuracy in target domains.
4) The improvement our approach yields in 1-shot tasks is more pronounced than in 5-shot tasks. We attribute this phenomenon to the limited knowledge contained within the smaller sample sizes of 1-shot tasks, which necessitates a greater reliance on the generalization capabilities and experiential knowledge provided by large models and textual modalities.
5) Our method's performance on the ChestX dataset is generally lower than that of other approaches. We hypothesize that this is due to the CLIP pre-trained model's limited exposure to relevant images during its pre-training phase, resulting in inferior performance for both CLIP and our CLIP backbone-based SeGD-VPT method.

\begin{table*}[ht]
\centering
\begin{tabular}{lcccccccc}
\hline
\textbf{1-shot} & \textbf{Backbone} & \textbf{Source} & \textbf{FT} & \textbf{ChestX} & \textbf{ISIC} & \textbf{EuroSAT} & \textbf{CropDisease} &\textbf{AVG}\\ \hline
GNN~\cite{gnn}        & RN10              &Y& -            & 22.00±0.46    & 32.02±0.66     & 63.69±1.03        &64.48±1.08  &45.55 \\
FWT~\cite{FWT}        & RN10              &Y& -            & 22.04±0.44    & 31.58±0.67     & 62.36±1.05        &66.36±1.04  &45.59  \\
LRP~\cite{sun2021explanation}        & RN10              &Y& -            &22.11±0.20     &30.94±0.30      &54.99±0.50         &59.23±0.50  &41.82  \\
\fyq{ATA}~\cite{wang2021cross} & RN10 & Y & - & 22.10±0.20 & 33.21±0.40 & 61.35±0.50 & 67.47±0.50 &46.03 \\
AFA~\cite{hu2022adversarial}        & RN10              &Y &-            &22.92±0.20     &33.21±0.30      &63.12±0.50         &67.61±0.50  &46.72   \\
wave-SAN~\cite{fu2022wave}   & RN10              &Y&-            &22.93±0.49     &33.35±0.71      &69.64±1.09         &70.80±1.06  &49.18 \\
\fyq{StyleAdv}~\cite{fu2023styleadv} & RN10   & Y & - &    22.64±0.35 &	33.96±0.57	& 70.94±0.82 &	74.13±0.78 &50.42
\\\hline
ATA-FT~\cite{wang2021cross}     &RN10               &Y&Y             &22.15±0.20     &34.94±0.40      &68.62±0.50         &75.41±0.50  &50.28   \\
\fyq{DARA}~\cite{zhao2023dual} & RN10 & Y & Y &22.92±0.40& 36.42±0.64 &	67.42±0.8 & 80.74±0.76 &51.88 \\
\fyq{StyleAdv-FT}~\cite{fu2023styleadv} & RN10    & Y & Y & 22.64±0.35 & 35.76±0.52 & 72.92±0.75	& 80.69±0.28 &53.00    
\\\hline
PMF*~\cite{hu2022pushing}       &ViT/DINO           &Y&Y             &21.73±0.30     &30.36±0.36      &70.74±0.63         &80.79±0.62  &50.91   \\
StyleAdv-FT~\cite{fu2023styleadv} &ViT/DINO           &Y&Y             &22.92±0.32     &33.99±0.46      &74.93±0.58         &84.11±0.57  &53.99   \\
\hline
FN+VDB~\cite{yazdanpanah2022visual}     &RN18               &-&Y             &22.64±0.41     &32.96±0.57      &69.67±0.80         &79.68±0.74  &51.24 \\
\fyq{IM-DCL~\cite{xu2024enhancing}} &RN10&-&Y& \textbf{23.98±0.79} &\textbf{	38.13±0.57} & 	77.14±0.71 & 	84.37±0.99 & 55.91\\
\textbf{SeGD-VPT (Ours)}   &ViT/CLIP     &-&Y             &22.03±0.32     &37.18±0.50&\textbf{83.58±0.52}&\textbf{90.45±0.52}
&\textbf{58.31}\\ \hline
\hline
\textbf{5-shot} & \textbf{Backbone} & \textbf{Source} & \textbf{FT} & \textbf{ChestX} & \textbf{ISIC} & \textbf{EuroSAT} & \textbf{CropDisease} &\textbf{AVG}\\ \hline
GNN~\cite{gnn}        & RN10              &Y& -           &25.27±0.46      &43.94±0.67      &83.64±0.77         &87.96±0.67 &60.20   \\
FWT~\cite{FWT}        & RN10              &Y & -          &25.18±0.45      &43.17±0.70      &83.01±0.79         &87.11±0.67 &59.62   \\
LRP~\cite{sun2021explanation}         & RN10              &Y& -           &24.53±0.30      &44.14±0.40      &77.14±0.40         &86.15±0.40 &57.99  \\
\fyq{ATA}~\cite{wang2021cross} & RN10 & Y & - & 24.32±0.40 & 44.91±0.40 & 83.75±0.40 & 90.59±0.30 &60.89 \\
AFA~\cite{hu2022adversarial}        & RN10              &Y&-            &25.02±0.20      &46.01±0.40      &85.58±0.40         &88.06±0.30 &61.17 \\
wave-SAN~\cite{fu2022wave}   & RN10              &Y&-            &25.63±0.49      &44.93±0.67      &85.22±0.71         &89.70±0.64 &61.37   \\
\fyq{StyleAdv}~\cite{fu2023styleadv} & RN10 & Y & - & 26.07±0.37& 45.77±0.51 & 86.58±0.54	& 93.65±0.39 &63.02
\\\hline
Fine-tune~\cite{guo2020broader}  &RN10               &Y&Y            &25.97±0.41      &48.11±0.64      &79.08±0.61         &89.25±0.51 &60.60\\
NSAE~\cite{liang2021boosting}       &RN10               &Y&Y            &27.10±0.44      &54.05±0.63      &83.96±0.57         &93.14±0.47 &64.56\\
BSR~\cite{liu2020feature}        &RN10               &Y&Y            &26.84±0.44      &\textbf{54.42±0.66}      &80.89±0.61         &92.17±0.45  &63.58\\ 
ATA-FT~\cite{wang2021cross}     &RN10               &Y&Y            &25.08±0.20      &49.79±0.40      &89.64±0.30         &95.44±0.20 &64.99    \\
\fyq{DARA}~\cite{zhao2023dual} & RN10 & Y & Y &  27.54±0.42 & 56.28±0.66	& 85.84±0.54	& 95.32±0.34 &66.25\\
\fyq{StyleAdv-FT}~\cite{fu2023styleadv} & RN10 & Y & Y & 26.24±0.35 &	53.05±0.54 & 	91.64±0.43 & 	96.51±0.28&66.86 \\
\hline
PMF*~\cite{hu2022pushing}       &ViT/DINO           &Y&Y            &27.27           &50.12           &85.98              &92.96 &64.08   \\
StyleAdv-FT~\cite{fu2023styleadv} &ViT/DINO           &Y&Y            &26.97±0.33      &51.23±0.51      &90.12±0.33         &95.99±0.27 &66.08    \\\hline
FN+VDB~\cite{yazdanpanah2022visual}     &RN18               &-&Y            &25.55±0.43    &47.48±0.59      &87.31±0.50         &94.63±0.37 &64.74 \\
\fyq{IM-DCL~\cite{xu2024enhancing}} & RN10 & - & Y &  \textbf{28.93±0.41}	& 52.74±0.69 &	89.47±0.42	& 95.73±0.38 &66.72\\
\textbf{SeGD-VPT (Ours)}   &ViT/CLIP     &-&Y            &23.20±0.30      &53.10±0.51&\textbf{93.81±0.24}&\textbf{96.93±0.25} &   \textbf{66.76}\\
\hline
\end{tabular}
\caption{The accuracy(\%) of four target domain datasets under 5-way 1-shot and 5-way 5-shot tasks. Among all the competitors and baselines our SeGD-VPT framework achieves the best performance in most cases. \fyq{We use the "AVG" to denote the averaged results over four target datasets.}} 
\label{tab:main}
\vspace{-0.25in}
\end{table*}
\subsection{Ablation Study}
In this study, we conduct experiments in a 5-way 1-shot setting to evaluate the effectiveness of two key components of our framework: the transfer module and the semantic guidance module, which are further divided into the diversity prompt and the describe prompt modules. The study is designed in three parts and all baselines share the fixed-parameter CLIP backbone and apply traditional data augmentation (random rotation/flipping).
Firstly, each method directly fine-tunes on the target domain. For CLIP-based baseline, we attach an FC classifier, which is directly fine-tuned on target domain data for transfer.
Secondly, we introduce SeGD-b1 baseline to assess the effectiveness of utilized transfer modules i.e. Deep Prompt Tuning, Adapter and contrastive learning. The learning process is divided into two steps: training deep prompt tokens with contrastive learning to generate features, followed by classification using an FC classifier with adapters.
Thirdly, 
The experimental results are presented in Table \ref{ablation}.\par

From Table \ref{ablation}, we observe the following: 1) Comparing SeGD-VPT with both CLIP-base and SeGD-b1, it is evident that the transfer learning modules and the combination of diversity prompt and describe prompt contributes to the framework's performance. Notably, SeGD-VPT outperforms SeGD-b1 by 0.47\%, 0.61\%, 1.37\%, and 0.98\% across four benchmarks, respectively, while SeGD-b1 exhibits improvements over CLIP-base by margins of 0.31\%, 0.91\%, 8.03\%, and 2.42\% on the same benchmarks. 2) The substantial enhancements from SeGD-b1 compared to CLIP-base highlight the impact of incorporating transfer learning modules, demonstrating their effectiveness in adapting the CLIP model to the target domain with limited samples. 3) Further improvement is noted with SeGD-VPT over SeGD-b1, indicating the capability of our generative framework to produce effective features, thereby augmenting the efficiency of transfer learning. 4) The results of SeGD-b1 and SeGD-b2 are very similar, differing by only 0.07\%. The explanation provided in this paper is that although SeGD-b2 increased diversity by adding diversity-prompts, it did not guide the learning process with diversity semantic features. Consequently, the prompts did not generate additional semantic information. The randomness and increased parameters heightened the training difficulty, resulting in its average performance being similar to that of SeGD-b1.

\par
\begin{table*}
\centering
\begin{tabular}{cccccccccc}
\hline
    \textbf{Method}&   \textbf{FT} & \textbf{Trans-Learning}    &\textbf{Diversity-P} &\textbf{Describe-P} &\textbf{ChestX} &\textbf{ISIC} &\textbf{EuroSAT} &\textbf{CropDisease}&\textbf{AVG}\\ \hline
     CLIP-base&Y&-&-&-&21.25±0.26&35.66±0.49&74.18±0.59&87.05±0.57&54.54\\
     SeGD-b1&Y&Y&-&-&21.56±0.30&36.57±0.50&82.21±0.54&89.47±0.54&57.45\\
     SeGD-b2&Y&Y&Y&-&21.36±0.28&36.36±0.49&82.87±0.50&88.91±0.57&57.38\\
     SeGD-VPT&Y&Y&Y&Y&\textbf{22.03±0.32}&\textbf{37.18±0.50}&\textbf{83.58±0.52}&\textbf{90.45±0.52}&\textbf{58.31}\\\hline
\end{tabular}
\caption{Ablation study to verify the effectiveness of each component in SeGD-VPT framework. We report the results(\%) on ChestX, ISIC, EuroSAT and CropDisease benchmarks under the 5-way 1-shot task. "FT": Fine Tuning, "-P": Prompt, "Trans-": Transfer.}
 \vspace{-0.1in}
\label{ablation}
\end{table*}

\subsection{Analysis}
\noindent\textbf{One Step VS Two Steps}.\par
The SeGD-VPT introduced in this paper is a two-step training framework that offers two advantages over the traditional one-step training framework. Firstly, the features generated by our framework can be used as a data augmentation method applicable to any downstream architecture, thereby enhancing its performance. More importantly, the optimization objectives of our framework's two stages are contradictory: the first stage aims to increase diversity through diversity loss and random selection, while the second (classifying) stage seeks consistency. Therefore, employing a one-step approach would mix these divergent optimization directions and diminish the training effectiveness. To validate this perspective, we conducted an experiment with a one-step training approach, merging the two-step processes into a singular framework and directly training the modules in step one using additional classification loss, with all other aspects remaining identical to SeGD-VPT. The experimental results are presented in the Table \ref{twostep}.\par
From the table, it is observed that the one-step framework exhibits a significant decrease in accuracy compared to the two-step framework employed in SeGD-VPT, with declines of 1.8\%, 17.51\%, 21.86\%, and 20.08\% across the four datasets, respectively. This supports our hypothesis that the one-step method, by accommodating two opposing optimization directions simultaneously, leads to reduced training efficiency. Therefore, the adoption of a two-step training approach in the SeGD-VPT framework is validated as being reasonable.\par
\begin{table*}
    \centering
    
    \begin{tabular}{cccccc}
    \hline
        \textbf{Dataset} & \textbf{ChestX} & \textbf{ISIC} & \textbf{EuroSAT} & \textbf{CropDisease} &\textbf{AVG} \\\hline
        One-step & 21.40±0.25 & 35.59±0.44	 & 71.95±0.52 & 76.85±0.66 &51.45 \\
        Two-step (SeGD-VPT) &\textbf{23.20±0.30}&\textbf{53.10±0.51}&\textbf{93.81±0.24}&\textbf{96.93±0.25}& \textbf{66.76}\\\hline
    \end{tabular}
    \caption{Analysis study to verified the Two-step training framework employed in SeGD-VPT is reasonable. We report the results(\%) on ChestX, ISIC, EuroSAT and CropDisease benchmarks under the 5-way 5-shot task.}
    \label{twostep}
\end{table*}
\noindent\textbf{Is Diversity Important?}\par
Our paper introduces a two-fold strategy to infuse diversity into augmentation samples. Firstly, we employ a diversity loss to prevent the convergence of learning across different prompts. Secondly, we enhance the learning process's diversity by randomly selecting texts for contrastive learning. Our work underscores the significance of incorporating diversity, as demonstrated through comparative experiments against two baselines, namely SeGD-b3 and SeGD-b4. For SeGD-b3, both aforementioned two modules are omitted; concretely, diversity loss is removed, and instead of random selection, the top 100 similar text features are directly used for contrastive learning. SeGD-b4 builds on \fyq{SeGD-b3} by reintroducing the module for random text selection. The experimental results are detailed in Table \ref{tab:diversity}. \par
The table substantiates the following points: 1) Both modules are effective, as evidenced by the average accuracy rates, with SeGD-VPT surpassing SeGD-b4 and SeGD-b3 by 0.13\% and 0.37\%, respectively. Moreover, SeGD-VPT achieved the best results in three datasets, except for a slight underperformance on the ISIC dataset by 0.05\% compared to SeGD-b4. We explaination that the introduction of random text selection has introduced a degree of randomness into the learning process, and a 0.05\% difference is not substantial, rendering this outcome acceptable. 2) A comparison reveals that SeGD-b4 improved upon SeGD-b3 by 0.24\%, roughly twice the improvement of SeGD-VPT over SeGD-b4. Our interpretation is that SeGD-b3, by eliminating both the $L_{div}$ and random text selection, completely removes randomness, leading to convergence in learning across different prompts and reducing the diversity of the generated features, hence diminishing generation efficiency. SeGD-b4, while also eliminating the loss, retains random text selection, allowing different prompts to be guided by different texts, thus enhancing diversity to a certain extent. Therefore, the improvement of SeGD-b4 over SeGD-b3 is more pronounced.

\begin{table*}
    \centering
    \begin{tabular}{cccccccc}
    \hline
        \textbf{Method} & \textbf{Prompt-D}& \textbf{Description-D}& \textbf{ChestX} & \textbf{ISIC} & \textbf{EuroSAT} & \textbf{CropDisease} &\textbf{AVG} \\\hline
        SeGD-b3 &-&-& 22.83±0.29&52.62±0.49&93.47±0.25&96.62±0.27&66.39\\
        SeGD-b4 &Y&-&22.96±0.29&	\textbf{53.15±0.50}&	93.63±0.25&	96.77±0.27&66.63\\
        SeGD-VPT &Y&Y&\textbf{23.20±0.30}&53.10±0.51&\textbf{93.81±0.24}&\textbf{96.93±0.25}& \textbf{66.76} \\\hline
    \end{tabular}
    \caption{Analysis study of the impact of the $L_{div}$ and random text selection on accuracy(\%) cross four benchmarks. "Prompt-D" is the abbreviation of "Prompt Diversity" means using $L_{div}$, "Description-D" is the abbreviation of "Description Diversity" means using random text selection.}
    \label{tab:diversity}
\end{table*}
\noindent\textbf{The Number of Diversity Prompts.}\par
\begin{figure}[h]
  \centering
  \includegraphics[width=\linewidth]{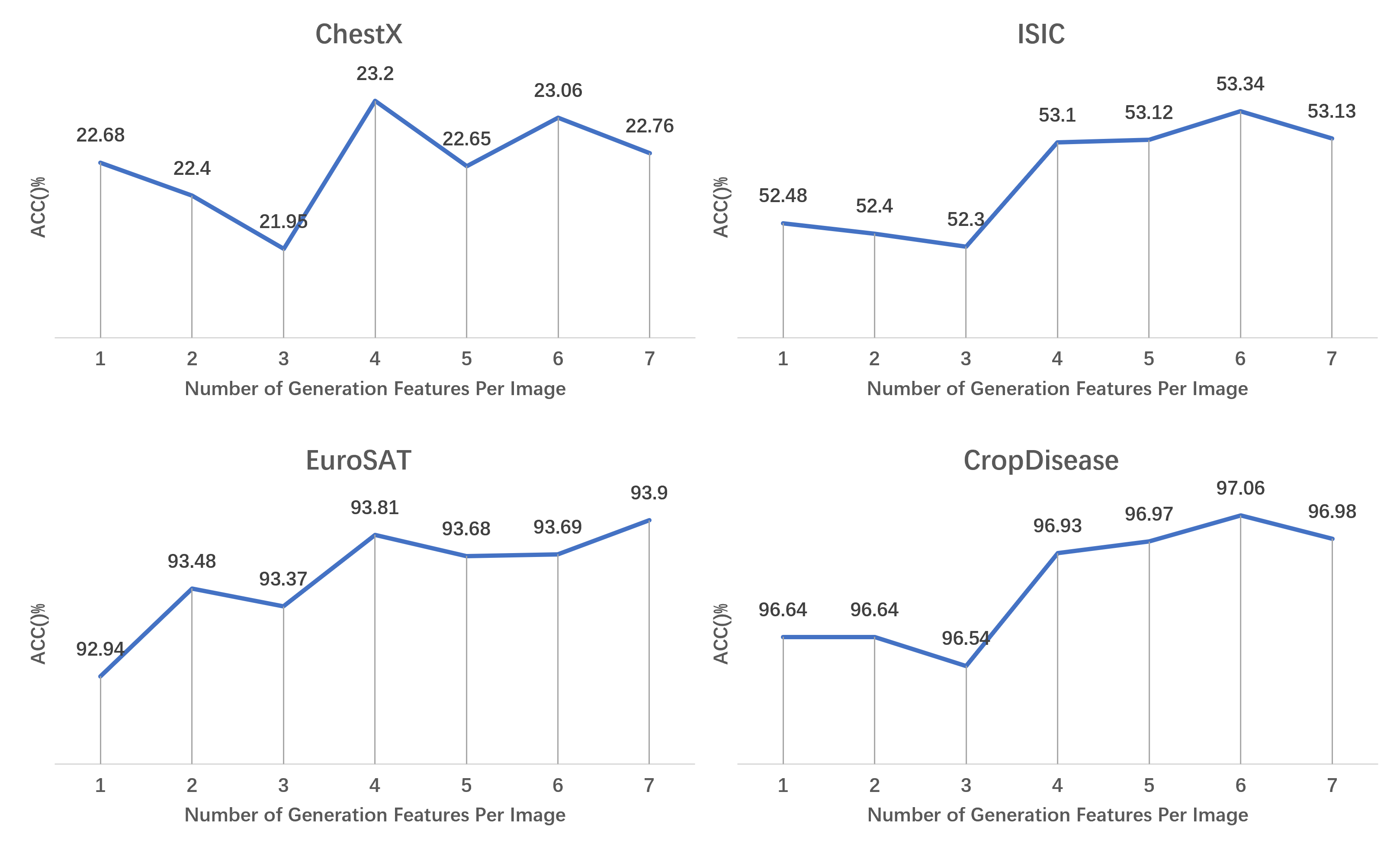}
  \caption{Accuracy curves demonstrate the impact of different feature counts per support sample across ChestX, ISIC, EuroSAT, and CropDisease datasets Under 5-way 5-shot conditions. Y-axis: Accuracy(\%); X-axis: Number of generation features per image (support sample).}
   \vspace{-0.1in}
  \label{fig:length}
\end{figure}

We explore the effect of generating varying numbers of features on accuracy. Specifically, We conduct experiments to generate from 1 to 7 features for each support sample in 5-shot task and illustrate the resulting variations in Figure \ref{fig:length}.\par
From Figure \ref{fig:length}, it is evident that the trend of accuracy changes across four datasets with the variation in the number of generated features follows a similar pattern, which can be summarized into two phases. Initially, when the number of generated features is relatively low (points 1-4), the accuracy increases more rapidly. Subsequently, the trend of increase becomes more gradual (points 4-7). Our explanation is that, at first, when there are fewer samples, the algorithm framework extracts less information from the texts, leaving more effective information unused, leading to a faster increase in accuracy as the number of generated samples rises. As the number of generated samples gradually increases, the remaining useful information in the texts becomes scarcer, and the addition of generated features makes the training more challenging, thus slowing the increase and, in some cases, such as with the ChestX dataset, leading to slight fluctuations.\par

\begin{figure*}[ht]
  \centering
   \vspace{-0.05in}
  \includegraphics[width=\linewidth]{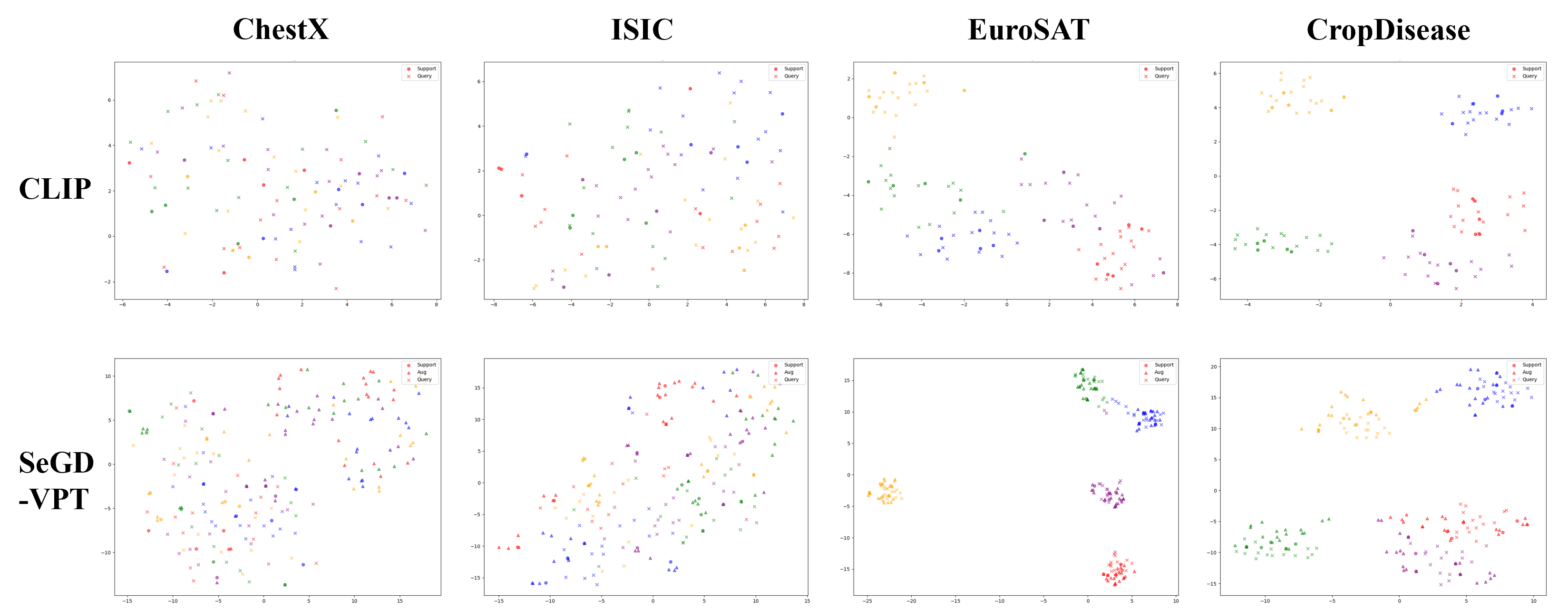}
  \caption{The t-SNE visualization results of our SeGD-VPT and CLIP model under 5-way 5-shot task cross four benchmarks. Different color represents different class in an episode, while different shapes, namely $\odot$, $\triangle$, and $\times$, represent support sample features, generated features, and query sample features, respectively.}
  \label{fig:visulization}
   \vspace{-0.05in}
\end{figure*}
\section{Visualization}
To demonstrate the effectiveness of the proposed algorithm in enhancing data distribution diversity, we conducted visualization experiments. Specifically, we carried out experiments for 5-shot task. For each experiment, we randomly selected five classes from four datasets and trained them using the step one process within the SeGD-VPT framework. Upon completion of the training, we projected support samples features, generated features and query samples features into a 2-D space using the t-SNE algorithm and displayed the results in Figure \ref{fig:visulization}. We also visualize the original CLIP features for comparison.\par
From Figure \ref{fig:visulization}, it can be observed that: 1) Features generated through SeGD-VPT are situated within the class distributions, indicating that the generated features can effectively represent samples of the relevant classes. 2) The generation of features effectively expands the distribution of samples. 3) Compared to features generated by CLIP, those produced by SeGD-VPT are more compact within the same class and have greater inter-class distances, which is more conducive to class discrimination. And this also indicates that the feature generation process can effectively transfer the model to the target domain.\par

\section{Conclusion}

In conclusion, this study introduces the SeGD-VPT framework for the source-free cross-domain few-shot learning (CD-FSL) task, successfully transferring pretrained models to target domains with minimal samples. By generating prompt visual features under the guidance of semantic modality to increase input diversity, and by implementing deep prompt tuning, our approach significantly enhances both transfer efficiency and model adaptability. Extensive experiments across various benchmarks have demonstrated that our framework not only rivals state-of-the-art models relying on source data but also sets a new standard in the source-free CD-FSL setting. These findings underscore the potential of leveraging textual information and innovative training strategies in overcoming the challenges of few-shot learning across domain gaps.

\clearpage

\bibliographystyle{ACM-Reference-Format}
\bibliography{sample-base}


\begin{thebibliography}{55}


\ifx \showCODEN    \undefined \def \showCODEN     #1{\unskip}     \fi
\ifx \showDOI      \undefined \def \showDOI       #1{#1}\fi
\ifx \showISBNx    \undefined \def \showISBNx     #1{\unskip}     \fi
\ifx \showISBNxiii \undefined \def \showISBNxiii  #1{\unskip}     \fi
\ifx \showISSN     \undefined \def \showISSN      #1{\unskip}     \fi
\ifx \showLCCN     \undefined \def \showLCCN      #1{\unskip}     \fi
\ifx \shownote     \undefined \def \shownote      #1{#1}          \fi
\ifx \showarticletitle \undefined \def \showarticletitle #1{#1}   \fi
\ifx \showURL      \undefined \def \showURL       {\relax}        \fi
\providecommand\bibfield[2]{#2}
\providecommand\bibinfo[2]{#2}
\providecommand\natexlab[1]{#1}
\providecommand\showeprint[2][]{arXiv:#2}

\bibitem[Auty and Mikolajczyk(2023)]%
        {auty2023learning}
\bibfield{author}{\bibinfo{person}{Dylan Auty} {and} \bibinfo{person}{Krystian Mikolajczyk}.} \bibinfo{year}{2023}\natexlab{}.
\newblock \showarticletitle{Learning to Prompt CLIP for Monocular Depth Estimation: Exploring the Limits of Human Language}. In \bibinfo{booktitle}{\emph{Proceedings of the IEEE/CVF International Conference on Computer Vision}}. \bibinfo{pages}{2039--2047}.
\newblock


\bibitem[Bose et~al\mbox{.}(2024)]%
        {bose2024stylip}
\bibfield{author}{\bibinfo{person}{Shirsha Bose}, \bibinfo{person}{Ankit Jha}, \bibinfo{person}{Enrico Fini}, \bibinfo{person}{Mainak Singha}, \bibinfo{person}{Elisa Ricci}, {and} \bibinfo{person}{Biplab Banerjee}.} \bibinfo{year}{2024}\natexlab{}.
\newblock \showarticletitle{Stylip: Multi-scale style-conditioned prompt learning for clip-based domain generalization}. In \bibinfo{booktitle}{\emph{Proceedings of the IEEE/CVF Winter Conference on Applications of Computer Vision}}. \bibinfo{pages}{5542--5552}.
\newblock


\bibitem[Chen et~al\mbox{.}(2023)]%
        {chen2023knowledge}
\bibfield{author}{\bibinfo{person}{Jingjing Chen}, \bibinfo{person}{Linhai Zhuo}, \bibinfo{person}{Zhipeng Wei}, \bibinfo{person}{Hao Zhang}, \bibinfo{person}{Huazhu Fu}, {and} \bibinfo{person}{Yu-Gang Jiang}.} \bibinfo{year}{2023}\natexlab{}.
\newblock \showarticletitle{Knowledge driven weights estimation for large-scale few-shot image recognition}.
\newblock \bibinfo{journal}{\emph{Pattern Recognition}}  \bibinfo{volume}{142} (\bibinfo{year}{2023}), \bibinfo{pages}{109668}.
\newblock


\bibitem[Cho et~al\mbox{.}(2023)]%
        {cho2023promptstyler}
\bibfield{author}{\bibinfo{person}{Junhyeong Cho}, \bibinfo{person}{Gilhyun Nam}, \bibinfo{person}{Sungyeon Kim}, \bibinfo{person}{Hunmin Yang}, {and} \bibinfo{person}{Suha Kwak}.} \bibinfo{year}{2023}\natexlab{}.
\newblock \showarticletitle{Promptstyler: Prompt-driven style generation for source-free domain generalization}. In \bibinfo{booktitle}{\emph{Proceedings of the IEEE/CVF International Conference on Computer Vision}}. \bibinfo{pages}{15702--15712}.
\newblock


\bibitem[Codella et~al\mbox{.}(2019)]%
        {codella2019skin}
\bibfield{author}{\bibinfo{person}{Noel Codella}, \bibinfo{person}{Veronica Rotemberg}, \bibinfo{person}{Philipp Tschandl}, \bibinfo{person}{M~Emre Celebi}, \bibinfo{person}{Stephen Dusza}, \bibinfo{person}{David Gutman}, \bibinfo{person}{Brian Helba}, \bibinfo{person}{Aadi Kalloo}, \bibinfo{person}{Konstantinos Liopyris}, \bibinfo{person}{Michael Marchetti}, {et~al\mbox{.}}} \bibinfo{year}{2019}\natexlab{}.
\newblock \showarticletitle{Skin lesion analysis toward melanoma detection 2018: A challenge hosted by the international skin imaging collaboration (isic)}.
\newblock \bibinfo{journal}{\emph{arXiv preprint arXiv:1902.03368}} (\bibinfo{year}{2019}).
\newblock


\bibitem[Deng et~al\mbox{.}(2019)]%
        {deng2019arcface}
\bibfield{author}{\bibinfo{person}{Jiankang Deng}, \bibinfo{person}{Jia Guo}, \bibinfo{person}{Niannan Xue}, {and} \bibinfo{person}{Stefanos Zafeiriou}.} \bibinfo{year}{2019}\natexlab{}.
\newblock \showarticletitle{Arcface: Additive angular margin loss for deep face recognition}. In \bibinfo{booktitle}{\emph{Proceedings of the IEEE/CVF conference on computer vision and pattern recognition}}. \bibinfo{pages}{4690--4699}.
\newblock


\bibitem[Fu et~al\mbox{.}(2021)]%
        {metafu}
\bibfield{author}{\bibinfo{person}{Yuqian Fu}, \bibinfo{person}{Yanwei Fu}, {and} \bibinfo{person}{Yu-Gang Jiang}.} \bibinfo{year}{2021}\natexlab{}.
\newblock \showarticletitle{Meta-FDMixup: Cross-Domain Few-Shot Learning Guided by Labeled Target Data}. In \bibinfo{booktitle}{\emph{Proceedings of the 29th ACM International Conference on Multimedia}}. \bibinfo{pages}{5326--5334}.
\newblock


\bibitem[Fu et~al\mbox{.}(2022a)]%
        {fu2022me}
\bibfield{author}{\bibinfo{person}{Yuqian Fu}, \bibinfo{person}{Yu Xie}, \bibinfo{person}{Yanwei Fu}, \bibinfo{person}{Jingjing Chen}, {and} \bibinfo{person}{Yu-Gang Jiang}.} \bibinfo{year}{2022}\natexlab{a}.
\newblock \showarticletitle{ME-D2N: Multi-Expert Domain Decompositional Network for Cross-Domain Few-Shot Learning}. In \bibinfo{booktitle}{\emph{Proceedings of the 30th ACM International Conference on Multimedia}}. \bibinfo{pages}{6609--6617}.
\newblock


\bibitem[Fu et~al\mbox{.}(2022b)]%
        {fu2022wave}
\bibfield{author}{\bibinfo{person}{Yuqian Fu}, \bibinfo{person}{Yu Xie}, \bibinfo{person}{Yanwei Fu}, \bibinfo{person}{Jingjing Chen}, {and} \bibinfo{person}{Yu-Gang Jiang}.} \bibinfo{year}{2022}\natexlab{b}.
\newblock \showarticletitle{Wave-SAN: Wavelet based Style Augmentation Network for Cross-Domain Few-Shot Learning}.
\newblock \bibinfo{journal}{\emph{arXiv preprint arXiv:2203.07656}} (\bibinfo{year}{2022}).
\newblock


\bibitem[Fu et~al\mbox{.}(2023)]%
        {fu2023styleadv}
\bibfield{author}{\bibinfo{person}{Yuqian Fu}, \bibinfo{person}{Yu Xie}, \bibinfo{person}{Yanwei Fu}, {and} \bibinfo{person}{Yu-Gang Jiang}.} \bibinfo{year}{2023}\natexlab{}.
\newblock \showarticletitle{StyleAdv: Meta Style Adversarial Training for Cross-Domain Few-Shot Learning}. In \bibinfo{booktitle}{\emph{Proceedings of the IEEE/CVF Conference on Computer Vision and Pattern Recognition}}. \bibinfo{pages}{24575--24584}.
\newblock


\bibitem[Gao et~al\mbox{.}(2021)]%
        {gao2021clip}
\bibfield{author}{\bibinfo{person}{Peng Gao}, \bibinfo{person}{Shijie Geng}, \bibinfo{person}{Renrui Zhang}, \bibinfo{person}{Teli Ma}, \bibinfo{person}{Rongyao Fang}, \bibinfo{person}{Yongfeng Zhang}, \bibinfo{person}{Hongsheng Li}, {and} \bibinfo{person}{Yu Qiao}.} \bibinfo{year}{2021}\natexlab{}.
\newblock \showarticletitle{Clip-adapter: Better vision-language models with feature adapters}.
\newblock \bibinfo{journal}{\emph{arXiv preprint arXiv:2110.04544}} (\bibinfo{year}{2021}).
\newblock


\bibitem[Garcia and Bruna(2017)]%
        {gnn}
\bibfield{author}{\bibinfo{person}{Victor Garcia} {and} \bibinfo{person}{Joan Bruna}.} \bibinfo{year}{2017}\natexlab{}.
\newblock \showarticletitle{Few-shot learning with graph neural networks}.
\newblock \bibinfo{journal}{\emph{arXiv preprint arXiv:1711.04043}} (\bibinfo{year}{2017}).
\newblock


\bibitem[Goyal et~al\mbox{.}(2023)]%
        {goyal2023finetune}
\bibfield{author}{\bibinfo{person}{Sachin Goyal}, \bibinfo{person}{Ananya Kumar}, \bibinfo{person}{Sankalp Garg}, \bibinfo{person}{Zico Kolter}, {and} \bibinfo{person}{Aditi Raghunathan}.} \bibinfo{year}{2023}\natexlab{}.
\newblock \showarticletitle{Finetune like you pretrain: Improved finetuning of zero-shot vision models}. In \bibinfo{booktitle}{\emph{Proceedings of the IEEE/CVF Conference on Computer Vision and Pattern Recognition}}. \bibinfo{pages}{19338--19347}.
\newblock


\bibitem[Guo et~al\mbox{.}(2020)]%
        {guo2020broader}
\bibfield{author}{\bibinfo{person}{Yunhui Guo}, \bibinfo{person}{Noel~C Codella}, \bibinfo{person}{Leonid Karlinsky}, \bibinfo{person}{James~V Codella}, \bibinfo{person}{John~R Smith}, \bibinfo{person}{Kate Saenko}, \bibinfo{person}{Tajana Rosing}, {and} \bibinfo{person}{Rogerio Feris}.} \bibinfo{year}{2020}\natexlab{}.
\newblock \showarticletitle{A broader study of cross-domain few-shot learning}. In \bibinfo{booktitle}{\emph{Computer Vision--ECCV 2020: 16th European Conference, Glasgow, UK, August 23--28, 2020, Proceedings, Part XXVII 16}}. Springer, \bibinfo{pages}{124--141}.
\newblock


\bibitem[Helber et~al\mbox{.}(2019)]%
        {helber2019eurosat}
\bibfield{author}{\bibinfo{person}{Patrick Helber}, \bibinfo{person}{Benjamin Bischke}, \bibinfo{person}{Andreas Dengel}, {and} \bibinfo{person}{Damian Borth}.} \bibinfo{year}{2019}\natexlab{}.
\newblock \showarticletitle{Eurosat: A novel dataset and deep learning benchmark for land use and land cover classification}.
\newblock \bibinfo{journal}{\emph{IEEE Journal of Selected Topics in Applied Earth Observations and Remote Sensing}} \bibinfo{volume}{12}, \bibinfo{number}{7} (\bibinfo{year}{2019}), \bibinfo{pages}{2217--2226}.
\newblock


\bibitem[Hu et~al\mbox{.}(2022)]%
        {hu2022pushing}
\bibfield{author}{\bibinfo{person}{Shell~Xu Hu}, \bibinfo{person}{Da Li}, \bibinfo{person}{Jan St{\"u}hmer}, \bibinfo{person}{Minyoung Kim}, {and} \bibinfo{person}{Timothy~M Hospedales}.} \bibinfo{year}{2022}\natexlab{}.
\newblock \showarticletitle{Pushing the Limits of Simple Pipelines for Few-Shot Learning: External Data and Fine-Tuning Make a Difference}. In \bibinfo{booktitle}{\emph{Proceedings of the IEEE/CVF Conference on Computer Vision and Pattern Recognition}}. \bibinfo{pages}{9068--9077}.
\newblock


\bibitem[Hu and Ma(2022)]%
        {hu2022adversarial}
\bibfield{author}{\bibinfo{person}{Yanxu Hu} {and} \bibinfo{person}{Andy~J Ma}.} \bibinfo{year}{2022}\natexlab{}.
\newblock \showarticletitle{Adversarial Feature Augmentation for Cross-domain Few-shot Classification}. In \bibinfo{booktitle}{\emph{Proceedings of the European Conference on Computer Vision (ECCV)}}.
\newblock


\bibitem[Ilharco et~al\mbox{.}(2021)]%
        {ilharco_gabriel_2021_5143773}
\bibfield{author}{\bibinfo{person}{Gabriel Ilharco}, \bibinfo{person}{Mitchell Wortsman}, \bibinfo{person}{Ross Wightman}, \bibinfo{person}{Cade Gordon}, \bibinfo{person}{Nicholas Carlini}, \bibinfo{person}{Rohan Taori}, \bibinfo{person}{Achal Dave}, \bibinfo{person}{Vaishaal Shankar}, \bibinfo{person}{Hongseok Namkoong}, \bibinfo{person}{John Miller}, \bibinfo{person}{Hannaneh Hajishirzi}, \bibinfo{person}{Ali Farhadi}, {and} \bibinfo{person}{Ludwig Schmidt}.} \bibinfo{year}{2021}\natexlab{}.
\newblock \bibinfo{booktitle}{\emph{OpenCLIP}}.
\newblock
\urldef\tempurl%
\url{https://doi.org/10.5281/zenodo.5143773}
\showDOI{\tempurl}
\newblock
\shownote{If you use this software, please cite it as below.}.


\bibitem[Islam et~al\mbox{.}(2021)]%
        {ddn}
\bibfield{author}{\bibinfo{person}{Ashraful Islam}, \bibinfo{person}{Chun-Fu~Richard Chen}, \bibinfo{person}{Rameswar Panda}, \bibinfo{person}{Leonid Karlinsky}, \bibinfo{person}{Rogerio Feris}, {and} \bibinfo{person}{Richard~J Radke}.} \bibinfo{year}{2021}\natexlab{}.
\newblock \showarticletitle{Dynamic distillation network for cross-domain few-shot recognition with unlabeled data}.
\newblock \bibinfo{journal}{\emph{Advances in Neural Information Processing Systems}}  \bibinfo{volume}{34} (\bibinfo{year}{2021}), \bibinfo{pages}{3584--3595}.
\newblock


\bibitem[Lee et~al\mbox{.}(2023)]%
        {lee2023read}
\bibfield{author}{\bibinfo{person}{Dongjun Lee}, \bibinfo{person}{Seokwon Song}, \bibinfo{person}{Jihee Suh}, \bibinfo{person}{Joonmyeong Choi}, \bibinfo{person}{Sanghyeok Lee}, {and} \bibinfo{person}{Hyunwoo~J Kim}.} \bibinfo{year}{2023}\natexlab{}.
\newblock \showarticletitle{Read-only prompt optimization for vision-language few-shot learning}. In \bibinfo{booktitle}{\emph{Proceedings of the IEEE/CVF International Conference on Computer Vision}}. \bibinfo{pages}{1401--1411}.
\newblock


\bibitem[Lester et~al\mbox{.}(2021)]%
        {lester2021power}
\bibfield{author}{\bibinfo{person}{Brian Lester}, \bibinfo{person}{Rami Al-Rfou}, {and} \bibinfo{person}{Noah Constant}.} \bibinfo{year}{2021}\natexlab{}.
\newblock \showarticletitle{The power of scale for parameter-efficient prompt tuning}.
\newblock \bibinfo{journal}{\emph{arXiv preprint arXiv:2104.08691}} (\bibinfo{year}{2021}).
\newblock


\bibitem[Li et~al\mbox{.}(2022b)]%
        {li2022learning}
\bibfield{author}{\bibinfo{person}{Aodi Li}, \bibinfo{person}{Liansheng Zhuang}, \bibinfo{person}{Shuo Fan}, {and} \bibinfo{person}{Shafei Wang}.} \bibinfo{year}{2022}\natexlab{b}.
\newblock \showarticletitle{Learning common and specific visual prompts for domain generalization}. In \bibinfo{booktitle}{\emph{Proceedings of the Asian Conference on Computer Vision}}. \bibinfo{pages}{4260--4275}.
\newblock


\bibitem[Li et~al\mbox{.}(2022a)]%
        {li2022targeted}
\bibfield{author}{\bibinfo{person}{Tianhong Li}, \bibinfo{person}{Peng Cao}, \bibinfo{person}{Yuan Yuan}, \bibinfo{person}{Lijie Fan}, \bibinfo{person}{Yuzhe Yang}, \bibinfo{person}{Rogerio~S Feris}, \bibinfo{person}{Piotr Indyk}, {and} \bibinfo{person}{Dina Katabi}.} \bibinfo{year}{2022}\natexlab{a}.
\newblock \showarticletitle{Targeted supervised contrastive learning for long-tailed recognition}. In \bibinfo{booktitle}{\emph{Proceedings of the IEEE/CVF Conference on Computer Vision and Pattern Recognition}}. \bibinfo{pages}{6918--6928}.
\newblock


\bibitem[Li and Liang(2021)]%
        {li2021prefix}
\bibfield{author}{\bibinfo{person}{Xiang~Lisa Li} {and} \bibinfo{person}{Percy Liang}.} \bibinfo{year}{2021}\natexlab{}.
\newblock \showarticletitle{Prefix-tuning: Optimizing continuous prompts for generation}.
\newblock \bibinfo{journal}{\emph{arXiv preprint arXiv:2101.00190}} (\bibinfo{year}{2021}).
\newblock


\bibitem[Liang et~al\mbox{.}(2021)]%
        {liang2021boosting}
\bibfield{author}{\bibinfo{person}{Hanwen Liang}, \bibinfo{person}{Qiong Zhang}, \bibinfo{person}{Peng Dai}, {and} \bibinfo{person}{Juwei Lu}.} \bibinfo{year}{2021}\natexlab{}.
\newblock \showarticletitle{Boosting the Generalization Capability in Cross-Domain Few-shot Learning via Noise-enhanced Supervised Autoencoder}. In \bibinfo{booktitle}{\emph{Proceedings of the IEEE/CVF International Conference on Computer Vision}}. \bibinfo{pages}{9424--9434}.
\newblock


\bibitem[Liu et~al\mbox{.}(2020)]%
        {liu2020feature}
\bibfield{author}{\bibinfo{person}{Bingyu Liu}, \bibinfo{person}{Zhen Zhao}, \bibinfo{person}{Zhenpeng Li}, \bibinfo{person}{Jianan Jiang}, \bibinfo{person}{Yuhong Guo}, {and} \bibinfo{person}{Jieping Ye}.} \bibinfo{year}{2020}\natexlab{}.
\newblock \showarticletitle{Feature transformation ensemble model with batch spectral regularization for cross-domain few-shot classification}. In \bibinfo{booktitle}{\emph{arXiv preprint arXiv:2005.08463}}.
\newblock


\bibitem[Liu et~al\mbox{.}(2022)]%
        {liu2022prompt}
\bibfield{author}{\bibinfo{person}{Lingbo Liu}, \bibinfo{person}{Jianlong Chang}, \bibinfo{person}{Bruce~XB Yu}, \bibinfo{person}{Liang Lin}, \bibinfo{person}{Qi Tian}, {and} \bibinfo{person}{Chang-Wen Chen}.} \bibinfo{year}{2022}\natexlab{}.
\newblock \showarticletitle{Prompt-matched semantic segmentation}.
\newblock \bibinfo{journal}{\emph{arXiv preprint arXiv:2208.10159}} (\bibinfo{year}{2022}).
\newblock


\bibitem[Liu et~al\mbox{.}(2023b)]%
        {liu2023pre}
\bibfield{author}{\bibinfo{person}{Pengfei Liu}, \bibinfo{person}{Weizhe Yuan}, \bibinfo{person}{Jinlan Fu}, \bibinfo{person}{Zhengbao Jiang}, \bibinfo{person}{Hiroaki Hayashi}, {and} \bibinfo{person}{Graham Neubig}.} \bibinfo{year}{2023}\natexlab{b}.
\newblock \showarticletitle{Pre-train, prompt, and predict: A systematic survey of prompting methods in natural language processing}.
\newblock \bibinfo{journal}{\emph{Comput. Surveys}} \bibinfo{volume}{55}, \bibinfo{number}{9} (\bibinfo{year}{2023}), \bibinfo{pages}{1--35}.
\newblock


\bibitem[Liu et~al\mbox{.}(2021)]%
        {liu2021p}
\bibfield{author}{\bibinfo{person}{Xiao Liu}, \bibinfo{person}{Kaixuan Ji}, \bibinfo{person}{Yicheng Fu}, \bibinfo{person}{Weng~Lam Tam}, \bibinfo{person}{Zhengxiao Du}, \bibinfo{person}{Zhilin Yang}, {and} \bibinfo{person}{Jie Tang}.} \bibinfo{year}{2021}\natexlab{}.
\newblock \showarticletitle{P-tuning v2: Prompt tuning can be comparable to fine-tuning universally across scales and tasks}.
\newblock \bibinfo{journal}{\emph{arXiv preprint arXiv:2110.07602}} (\bibinfo{year}{2021}).
\newblock


\bibitem[Liu et~al\mbox{.}(2023a)]%
        {liu2023generating}
\bibfield{author}{\bibinfo{person}{Zuhao Liu}, \bibinfo{person}{Xiao-Ming Wu}, \bibinfo{person}{Dian Zheng}, \bibinfo{person}{Kun-Yu Lin}, {and} \bibinfo{person}{Wei-Shi Zheng}.} \bibinfo{year}{2023}\natexlab{a}.
\newblock \showarticletitle{Generating Anomalies for Video Anomaly Detection With Prompt-Based Feature Mapping}. In \bibinfo{booktitle}{\emph{Proceedings of the IEEE/CVF Conference on Computer Vision and Pattern Recognition}}. \bibinfo{pages}{24500--24510}.
\newblock


\bibitem[Ma et~al\mbox{.}(2023)]%
        {ma2023prod}
\bibfield{author}{\bibinfo{person}{Tianyi Ma}, \bibinfo{person}{Yifan Sun}, \bibinfo{person}{Zongxin Yang}, {and} \bibinfo{person}{Yi Yang}.} \bibinfo{year}{2023}\natexlab{}.
\newblock \showarticletitle{ProD: Prompting-To-Disentangle Domain Knowledge for Cross-Domain Few-Shot Image Classification}. In \bibinfo{booktitle}{\emph{Proceedings of the IEEE/CVF Conference on Computer Vision and Pattern Recognition}}. \bibinfo{pages}{19754--19763}.
\newblock


\bibitem[Miyai et~al\mbox{.}(2023)]%
        {miyai2023locoop}
\bibfield{author}{\bibinfo{person}{Atsuyuki Miyai}, \bibinfo{person}{Qing Yu}, \bibinfo{person}{Go Irie}, {and} \bibinfo{person}{Kiyoharu Aizawa}.} \bibinfo{year}{2023}\natexlab{}.
\newblock \showarticletitle{LoCoOp: Few-Shot Out-of-Distribution Detection via Prompt Learning}.
\newblock \bibinfo{journal}{\emph{arXiv preprint arXiv:2306.01293}} (\bibinfo{year}{2023}).
\newblock


\bibitem[Mohanty et~al\mbox{.}(2016)]%
        {mohanty2016using}
\bibfield{author}{\bibinfo{person}{Sharada~P Mohanty}, \bibinfo{person}{David~P Hughes}, {and} \bibinfo{person}{Marcel Salath{\'e}}.} \bibinfo{year}{2016}\natexlab{}.
\newblock \showarticletitle{Using deep learning for image-based plant disease detection}.
\newblock \bibinfo{journal}{\emph{Frontiers in plant science}}  \bibinfo{volume}{7} (\bibinfo{year}{2016}), \bibinfo{pages}{1419}.
\newblock


\bibitem[Phoo and Hariharan(2020)]%
        {startup}
\bibfield{author}{\bibinfo{person}{Cheng~Perng Phoo} {and} \bibinfo{person}{Bharath Hariharan}.} \bibinfo{year}{2020}\natexlab{}.
\newblock \showarticletitle{Self-training for few-shot transfer across extreme task differences}.
\newblock \bibinfo{journal}{\emph{arXiv preprint arXiv:2010.07734}} (\bibinfo{year}{2020}).
\newblock


\bibitem[Qian et~al\mbox{.}(2023)]%
        {qian2023decouple}
\bibfield{author}{\bibinfo{person}{Zi Qian}, \bibinfo{person}{Xin Wang}, \bibinfo{person}{Xuguang Duan}, \bibinfo{person}{Pengda Qin}, \bibinfo{person}{Yuhong Li}, {and} \bibinfo{person}{Wenwu Zhu}.} \bibinfo{year}{2023}\natexlab{}.
\newblock \showarticletitle{Decouple before interact: Multi-modal prompt learning for continual visual question answering}. In \bibinfo{booktitle}{\emph{Proceedings of the IEEE/CVF International Conference on Computer Vision}}. \bibinfo{pages}{2953--2962}.
\newblock


\bibitem[Rong et~al\mbox{.}(2023)]%
        {rong2023retrieval}
\bibfield{author}{\bibinfo{person}{Jintao Rong}, \bibinfo{person}{Hao Chen}, \bibinfo{person}{Tianxiao Chen}, \bibinfo{person}{Linlin Ou}, \bibinfo{person}{Xinyi Yu}, {and} \bibinfo{person}{Yifan Liu}.} \bibinfo{year}{2023}\natexlab{}.
\newblock \showarticletitle{Retrieval-Enhanced Visual Prompt Learning for Few-shot Classification}.
\newblock \bibinfo{journal}{\emph{arXiv preprint arXiv:2306.02243}} (\bibinfo{year}{2023}).
\newblock


\bibitem[Song et~al\mbox{.}(2023)]%
        {song2023vppt}
\bibfield{author}{\bibinfo{person}{Zhao Song}, \bibinfo{person}{Ke Yang}, \bibinfo{person}{Naiyang Guan}, \bibinfo{person}{Junjie Zhu}, \bibinfo{person}{Peng Qiao}, {and} \bibinfo{person}{Qingyong Hu}.} \bibinfo{year}{2023}\natexlab{}.
\newblock \showarticletitle{VPPT: Visual Pre-Trained Prompt Tuning Framework for Few-Shot Image Classification}. In \bibinfo{booktitle}{\emph{ICASSP 2023-2023 IEEE International Conference on Acoustics, Speech and Signal Processing (ICASSP)}}. IEEE, \bibinfo{pages}{1--5}.
\newblock


\bibitem[Sun et~al\mbox{.}(2021)]%
        {sun2021explanation}
\bibfield{author}{\bibinfo{person}{Jiamei Sun}, \bibinfo{person}{Sebastian Lapuschkin}, \bibinfo{person}{Wojciech Samek}, \bibinfo{person}{Yunqing Zhao}, \bibinfo{person}{Ngai-Man Cheung}, {and} \bibinfo{person}{Alexander Binder}.} \bibinfo{year}{2021}\natexlab{}.
\newblock \showarticletitle{Explanation-guided training for cross-domain few-shot classification}. In \bibinfo{booktitle}{\emph{2020 25th International Conference on Pattern Recognition (ICPR)}}. IEEE, \bibinfo{pages}{7609--7616}.
\newblock


\bibitem[Tschandl et~al\mbox{.}(2018)]%
        {tschandl2018ham10000}
\bibfield{author}{\bibinfo{person}{Philipp Tschandl}, \bibinfo{person}{Cliff Rosendahl}, {and} \bibinfo{person}{Harald Kittler}.} \bibinfo{year}{2018}\natexlab{}.
\newblock \showarticletitle{The HAM10000 dataset, a large collection of multi-source dermatoscopic images of common pigmented skin lesions}.
\newblock \bibinfo{journal}{\emph{Scientific data}} \bibinfo{volume}{5}, \bibinfo{number}{1} (\bibinfo{year}{2018}), \bibinfo{pages}{1--9}.
\newblock


\bibitem[Tseng et~al\mbox{.}(2020)]%
        {FWT}
\bibfield{author}{\bibinfo{person}{Hung-Yu Tseng}, \bibinfo{person}{Hsin-Ying Lee}, \bibinfo{person}{Jia-Bin Huang}, {and} \bibinfo{person}{Ming-Hsuan Yang}.} \bibinfo{year}{2020}\natexlab{}.
\newblock \showarticletitle{Cross-domain few-shot classification via learned feature-wise transformation}.
\newblock \bibinfo{journal}{\emph{arXiv preprint arXiv:2001.08735}} (\bibinfo{year}{2020}).
\newblock


\bibitem[Vinyals et~al\mbox{.}(2016)]%
        {vinyals2016matching}
\bibfield{author}{\bibinfo{person}{Oriol Vinyals}, \bibinfo{person}{Charles Blundell}, \bibinfo{person}{Timothy Lillicrap}, \bibinfo{person}{Daan Wierstra}, {et~al\mbox{.}}} \bibinfo{year}{2016}\natexlab{}.
\newblock \showarticletitle{Matching networks for one shot learning}. In \bibinfo{booktitle}{\emph{Proceedings of the 30th Conference on Neural Information Processing Systems (NIPS)}}. \bibinfo{pages}{3630--3638}.
\newblock


\bibitem[Wang and Deng(2021a)]%
        {wang2021}
\bibfield{author}{\bibinfo{person}{Haoqing Wang} {and} \bibinfo{person}{Zhi-Hong Deng}.} \bibinfo{year}{2021}\natexlab{a}.
\newblock \showarticletitle{Cross-domain few-shot classification via adversarial task augmentation}.
\newblock \bibinfo{journal}{\emph{arXiv preprint arXiv:2104.14385}} (\bibinfo{year}{2021}).
\newblock


\bibitem[Wang and Deng(2021b)]%
        {wang2021cross}
\bibfield{author}{\bibinfo{person}{Haoqing Wang} {and} \bibinfo{person}{Zhi-Hong Deng}.} \bibinfo{year}{2021}\natexlab{b}.
\newblock \showarticletitle{Cross-domain few-shot classification via adversarial task augmentation}.
\newblock \bibinfo{journal}{\emph{arXiv preprint arXiv:2104.14385}} (\bibinfo{year}{2021}).
\newblock


\bibitem[Wang et~al\mbox{.}(2023)]%
        {wang2023large}
\bibfield{author}{\bibinfo{person}{Xiao Wang}, \bibinfo{person}{Guangyao Chen}, \bibinfo{person}{Guangwu Qian}, \bibinfo{person}{Pengcheng Gao}, \bibinfo{person}{Xiao-Yong Wei}, \bibinfo{person}{Yaowei Wang}, \bibinfo{person}{Yonghong Tian}, {and} \bibinfo{person}{Wen Gao}.} \bibinfo{year}{2023}\natexlab{}.
\newblock \showarticletitle{Large-scale multi-modal pre-trained models: A comprehensive survey}.
\newblock \bibinfo{journal}{\emph{Machine Intelligence Research}} (\bibinfo{year}{2023}), \bibinfo{pages}{1--36}.
\newblock


\bibitem[Wang et~al\mbox{.}(2017)]%
        {wang2017chestx}
\bibfield{author}{\bibinfo{person}{Xiaosong Wang}, \bibinfo{person}{Yifan Peng}, \bibinfo{person}{Le Lu}, \bibinfo{person}{Zhiyong Lu}, \bibinfo{person}{Mohammadhadi Bagheri}, {and} \bibinfo{person}{Ronald~M Summers}.} \bibinfo{year}{2017}\natexlab{}.
\newblock \showarticletitle{Chestx-ray8: Hospital-scale chest x-ray database and benchmarks on weakly-supervised classification and localization of common thorax diseases}. In \bibinfo{booktitle}{\emph{Proceedings of the IEEE conference on computer vision and pattern recognition}}. \bibinfo{pages}{2097--2106}.
\newblock


\bibitem[Wu et~al\mbox{.}(2023)]%
        {wu2023few}
\bibfield{author}{\bibinfo{person}{Yong Wu}, \bibinfo{person}{Shekhor Chanda}, \bibinfo{person}{Mehrdad Hosseinzadeh}, \bibinfo{person}{Zhi Liu}, {and} \bibinfo{person}{Yang Wang}.} \bibinfo{year}{2023}\natexlab{}.
\newblock \showarticletitle{Few-Shot Learning of Compact Models via Task-Specific Meta Distillation}. In \bibinfo{booktitle}{\emph{Proceedings of the IEEE/CVF Winter Conference on Applications of Computer Vision}}. \bibinfo{pages}{6265--6274}.
\newblock


\bibitem[Xu et~al\mbox{.}(2024)]%
        {xu2024enhancing}
\bibfield{author}{\bibinfo{person}{Huali Xu}, \bibinfo{person}{Li Liu}, \bibinfo{person}{Shuaifeng Zhi}, \bibinfo{person}{Shaojing Fu}, \bibinfo{person}{Zhuo Su}, \bibinfo{person}{Ming-Ming Cheng}, {and} \bibinfo{person}{Yongxiang Liu}.} \bibinfo{year}{2024}\natexlab{}.
\newblock \showarticletitle{Enhancing Information Maximization with Distance-Aware Contrastive Learning for Source-Free Cross-Domain Few-Shot Learning}.
\newblock \bibinfo{journal}{\emph{IEEE Transactions on Image Processing}} (\bibinfo{year}{2024}).
\newblock


\bibitem[Yan et~al\mbox{.}(2023)]%
        {yan2023prompt}
\bibfield{author}{\bibinfo{person}{Liqi Yan}, \bibinfo{person}{Cheng Han}, \bibinfo{person}{Zenglin Xu}, \bibinfo{person}{Dongfang Liu}, {and} \bibinfo{person}{Qifan Wang}.} \bibinfo{year}{2023}\natexlab{}.
\newblock \showarticletitle{Prompt learns prompt: exploring knowledge-aware generative prompt collaboration for video captioning}. In \bibinfo{booktitle}{\emph{Proceedings of international joint conference on artificial intelligence (IJCAI)}}. \bibinfo{pages}{1622--1630}.
\newblock


\bibitem[Yazdanpanah and Moradi(2022)]%
        {yazdanpanah2022visual}
\bibfield{author}{\bibinfo{person}{Moslem Yazdanpanah} {and} \bibinfo{person}{Parham Moradi}.} \bibinfo{year}{2022}\natexlab{}.
\newblock \showarticletitle{Visual domain bridge: A source-free domain adaptation for cross-domain few-shot learning}. In \bibinfo{booktitle}{\emph{Proceedings of the IEEE/CVF Conference on Computer Vision and Pattern Recognition}}. \bibinfo{pages}{2868--2877}.
\newblock


\bibitem[Zhang et~al\mbox{.}(2021)]%
        {zhang2021tip}
\bibfield{author}{\bibinfo{person}{Renrui Zhang}, \bibinfo{person}{Rongyao Fang}, \bibinfo{person}{Wei Zhang}, \bibinfo{person}{Peng Gao}, \bibinfo{person}{Kunchang Li}, \bibinfo{person}{Jifeng Dai}, \bibinfo{person}{Yu Qiao}, {and} \bibinfo{person}{Hongsheng Li}.} \bibinfo{year}{2021}\natexlab{}.
\newblock \showarticletitle{Tip-adapter: Training-free clip-adapter for better vision-language modeling}.
\newblock \bibinfo{journal}{\emph{arXiv preprint arXiv:2111.03930}} (\bibinfo{year}{2021}).
\newblock


\bibitem[Zhang et~al\mbox{.}(2023b)]%
        {zhang2023prompt}
\bibfield{author}{\bibinfo{person}{Renrui Zhang}, \bibinfo{person}{Xiangfei Hu}, \bibinfo{person}{Bohao Li}, \bibinfo{person}{Siyuan Huang}, \bibinfo{person}{Hanqiu Deng}, \bibinfo{person}{Yu Qiao}, \bibinfo{person}{Peng Gao}, {and} \bibinfo{person}{Hongsheng Li}.} \bibinfo{year}{2023}\natexlab{b}.
\newblock \showarticletitle{Prompt, generate, then cache: Cascade of foundation models makes strong few-shot learners}. In \bibinfo{booktitle}{\emph{Proceedings of the IEEE/CVF Conference on Computer Vision and Pattern Recognition}}. \bibinfo{pages}{15211--15222}.
\newblock


\bibitem[Zhang et~al\mbox{.}(2023a)]%
        {zhang2023domain}
\bibfield{author}{\bibinfo{person}{Xin Zhang}, \bibinfo{person}{Shixiang~Shane Gu}, \bibinfo{person}{Yutaka Matsuo}, {and} \bibinfo{person}{Yusuke Iwasawa}.} \bibinfo{year}{2023}\natexlab{a}.
\newblock \showarticletitle{Domain prompt learning for efficiently adapting clip to unseen domains}.
\newblock \bibinfo{journal}{\emph{Transactions of the Japanese Society for Artificial Intelligence}} \bibinfo{volume}{38}, \bibinfo{number}{6} (\bibinfo{year}{2023}), \bibinfo{pages}{B--MC2\_1}.
\newblock


\bibitem[Zhao et~al\mbox{.}(2023)]%
        {zhao2023dual}
\bibfield{author}{\bibinfo{person}{Yifan Zhao}, \bibinfo{person}{Tong Zhang}, \bibinfo{person}{Jia Li}, {and} \bibinfo{person}{Yonghong Tian}.} \bibinfo{year}{2023}\natexlab{}.
\newblock \showarticletitle{Dual adaptive representation alignment for cross-domain few-shot learning}.
\newblock \bibinfo{journal}{\emph{IEEE Transactions on Pattern Analysis and Machine Intelligence}} (\bibinfo{year}{2023}).
\newblock


\bibitem[Zheng et~al\mbox{.}(2023)]%
        {zheng2023cross}
\bibfield{author}{\bibinfo{person}{Hao Zheng}, \bibinfo{person}{Runqi Wang}, \bibinfo{person}{Jianzhuang Liu}, {and} \bibinfo{person}{Asako Kanezaki}.} \bibinfo{year}{2023}\natexlab{}.
\newblock \showarticletitle{Cross-level distillation and feature denoising for cross-domain few-shot classification}.
\newblock \bibinfo{journal}{\emph{arXiv preprint arXiv:2311.02392}} (\bibinfo{year}{2023}).
\newblock


\bibitem[Zhuo et~al\mbox{.}(2022)]%
        {zhuo2022tgdm}
\bibfield{author}{\bibinfo{person}{Linhai Zhuo}, \bibinfo{person}{Yuqian Fu}, \bibinfo{person}{Jingjing Chen}, \bibinfo{person}{Yixin Cao}, {and} \bibinfo{person}{Yu-Gang Jiang}.} \bibinfo{year}{2022}\natexlab{}.
\newblock \showarticletitle{TGDM: Target Guided Dynamic Mixup for Cross-Domain Few-Shot Learning}. In \bibinfo{booktitle}{\emph{Proceedings of the ACM International Conference on Multimedia (ACM MM)}}.
\newblock


\end{thebibliography}

\end{document}


\title{Supplementary Materials: SeGD-VPT: Semantic Guided Diversity Visual Prompt Tuning for Source Free CD-FSL}
\author{Anonymous Authors}

\maketitle
\section{Class Descriptions}
\begin{figure}[h]
    \centering
    \includegraphics[width=0.85\linewidth]{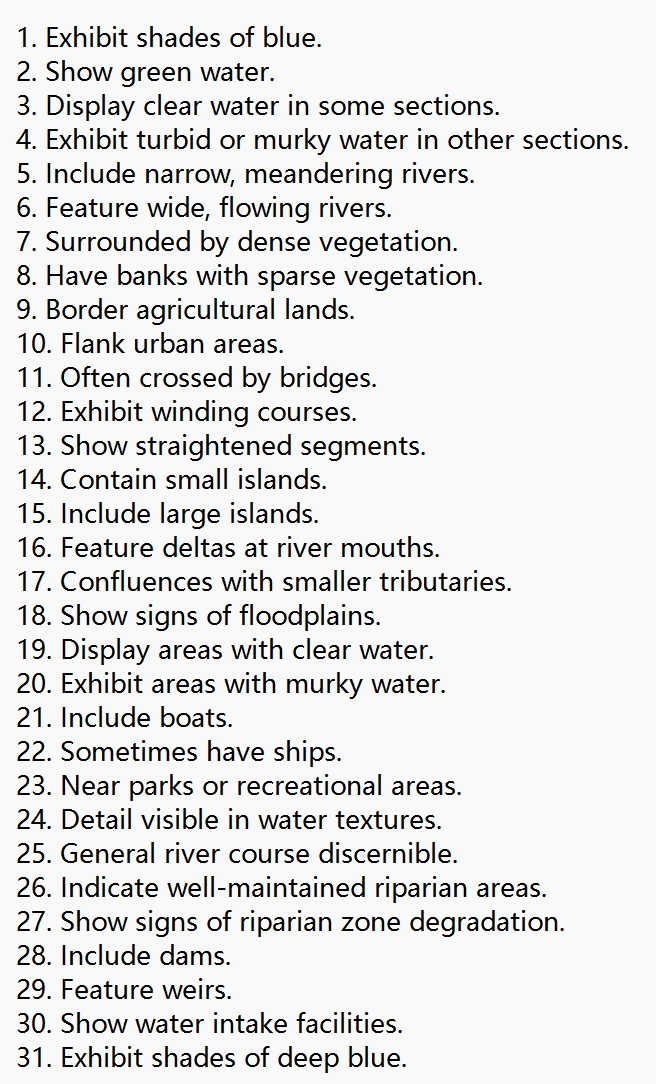} 
       \caption{Description examples from the "River" category in the EuroSAT dataset.}
    \label{fig:river}
\end{figure}

This supplementary material presents a sample from the "River" category in the EuroSAT dataset, as shown in Figure \ref{fig:river}. We use a predefined template "[Domain] photo of [Class] [Description]." to convert specific descriptions into input. For example, for the first description "Exhibit shades of blue", this input becomes "Satellite image photo of river exhibit shades of blue."

\newpage
\section{Training Details}
This supplementary material illustrates the training epochs $T$ and learning rates $lr$ for the four benchmarks across the 5-way 1-shot and 5-way 5-shot tasks, as detailed in Table \ref{tab:detail}.

\begin{table}[H]\small
    \centering
    \begin{tabular}{|c|cccc|}
    \hline
         1-shot&ChestX&ISIC&EuroSAT&CropDisease   \\\hline
         $T$ & 60& 60& 40 &60\\
         $lr$ &0.0001&0.0001&0.001&0.0001\\\hline
         5-shot&ChestX&ISIC&EuroSAT&CropDisease\\\hline
         $T$ & 60& 60& 55 &60\\
         $lr$ &0.0001&0.0001&0.001&0.0001\\\hline
    \end{tabular}
    \caption{Training epochs $T$ and learning rates $lr$ for the four benchmarks across the 5-way 1-shot and 5-way 5-shot tasks.}
    \label{tab:detail}
\end{table}